\documentclass[runningheads]{llncs}
\usepackage{graphicx}
\usepackage{placeins}
\usepackage{amsmath,amssymb}
\usepackage{relsize}
\usepackage{comment}
\usepackage{xspace}
\usepackage{booktabs}
\usepackage{afterpage}
\usepackage{algorithm}
\usepackage[noend]{algpseudocode}
\algnewcommand{\IfThenElse}[3]{%
  \State \algorithmicif\ #1\ \algorithmicthen\ #2\ \algorithmicelse\ #3}
\usepackage[square,numbers,sort]{natbib}
\makeatletter
\renewcommand\bibsection%
{
  \section*{\refname
    \@mkboth{\MakeUppercase{\refname}}{\MakeUppercase{\refname}}}
}
\renewcommand\@biblabel[1]{#1.}

\makeatother
\usepackage{hyperref}
\hypersetup{colorlinks,allcolors=blue}

\newcommand{\MakeBenchmarkName}[1]{
  \uppercase{\expandafter\newcommand\csname#1}\endcsname{%
    \textsc{#1}\xspace}}
\MakeBenchmarkName{yabbob}
\MakeBenchmarkName{yasmallbbob}
\MakeBenchmarkName{yatuningbbob}
\MakeBenchmarkName{yaparabbob}
\MakeBenchmarkName{yaboundedbbob}
\MakeBenchmarkName{yabigbbob} 
\MakeBenchmarkName{yaboxbbob}
\MakeBenchmarkName{yahdbbob}
\MakeBenchmarkName{yanoisyb}
\MakeBenchmarkName{yawidebbob}

\begin{document}
\title{Improving Nevergrad's Algorithm Selection Wizard NGOpt through Automated Algorithm Configuration} 
\titlerunning{Improving NGOpt through Automated Algorithm Configuration}
 \author{Risto Trajanov\inst{1}\orcidID{0000-0002-3604-2515}\and
 Ana Nikolikj\inst{2,3}\orcidID{0000-0002-6983-9627}\and
 Gjorgjina Cenikj\inst{2,3}\orcidID{0000-0002-2723-0821}\and
 Fabien Teytaud\inst{4}\and
 Mathurin Videau\inst{5}\and
 Olivier Teytaud\inst{5}\and
 Tome Eftimov\inst{2}\orcidID{0000-0001-7330-1902}\\
 \and
 Manuel L\'opez-Ib\'a\~nez\inst{6}\orcidID{0000-0001-9974-1295}\and
 Carola Doerr\inst{7}\orcidID{0000-0002-4981-3227}
}
\institute{
Faculty of Computer Science and Engineering, Ss. Cyril and Methodius Skopje, North Macedonia
\and
 Jožef Stefan Institute, Ljubljana, Slovenia\\
\and
Jozef Stefan International Postgraduate School, Ljubljana, Slovenia\and
Universit\'e Littoral C\^ote d'Opale\and
Facebook AI Research
\and
ITIS Software, Universidad de Málaga, Spain
\and
Sorbonne Universit\'e, CNRS, LIP6, Paris, France
}
\authorrunning{Trajanov, Nikolikj, Cenikj et al.}
\maketitle              %
\begin{abstract}
 Algorithm selection wizards are effective and versatile tools that automatically select an optimization algorithm given high-level information about the problem and available computational resources, such as number and type of decision variables, maximal number of evaluations, possibility to parallelize evaluations, etc. State-of-the-art algorithm selection wizards are complex and difficult to improve. We propose in this work the use of automated configuration methods for improving their performance by finding better configurations of the algorithms that compose them. In particular, we use elitist iterated racing (irace) to find CMA configurations for specific artificial benchmarks that replace the hand-crafted CMA configurations currently used in the NGOpt wizard provided by the Nevergrad platform. We discuss in detail the setup of irace for the purpose of generating configurations that work well over the diverse set of problem instances within each benchmark. Our approach improves the performance of the NGOpt wizard, even on benchmark suites that were not part of the tuning by irace.
\end{abstract}

\section{Introduction}
	
In the context of black-box optimization, the use of a portfolio of
optimization algorithms~\citep{Rice1976}, from which an algorithm is
selected depending on the features of the particular problem instance to be
solved, is becoming increasingly popular.  The algorithm that encapsulates the
selection rules and the portfolio of algorithms is sometimes referred as a
``wizard''~\cite{NGoptTEVC}. %
Algorithm selection wizards provide versatile, robust and convenient tools, particularly for
black-box optimization, where problems that arise in
practice show a large variety of different requirements, both with respect to problem models, but also with respect to the (computational) resources that are available to solve them. 

Building a competitive
algorithm selection wizard is not an easy task, because it not only
requires devising the rules for selecting an algorithm for a given 
problem instance, but also configuring the parameters of the algorithms to be
selected, which is a difficult task by 
itself~\cite{Birattari09tuning}. Although there are examples of wizards that
were build automatically for SAT problems~\cite{XuHutHooLey2008jair}, many
algorithm selection wizards are still hand-crafted. %
An example of a successful hand-crafted algorithm selection wizard is NGOpt~\cite{NGoptTEVC}, 
provided by the optimization platform Nevergrad~\cite{nevergrad}.

NGOpt was designed by carefully studying the performance of
tens of optimizers on a wide range of benchmark suites to design
hand-crafted rules that aim to select the best optimizer for particular
problem features. Through an iterative improvement process, new versions of
NGOpt are designed by refining the hand-crated rules and replacing optimizers
by others that lead to a better performance of the new NGOpt version. The
result is a complex algorithm selection wizard that outperforms many well-known stand-alone optimizers on a
wide range of benchmark suites.

In this work, we attempt to investigate the potential of improving NGOpt via automated algorithm design techniques.  
Rebuilding NGOpt from scratch using automated methods is a daunting task
that would waste the knowledge already encoded in it and the human effort
already invested in its creation. Instead, we examine how automatic
algorithm configuration (AC) methods may be used in a judicious manner to improve NGOpt by
focusing on improving one of its most critical
underlying optimizers.

AC methods such as irace~\cite{LopDubPerStuBir2016irace}, aim to find a configuration, i.e., a setting of the parameters of an algorithm, that gives good expected performance over a large space of problem instances by evaluating configurations on a \emph{training} subset from such space. AC methods, including irace, are typically designed to handle categorical and numerical parameter spaces and stochastic performance metrics, not only due to the inherent stochasticity of randomized optimization algorithms but also because the training instances represent a random sample of the problems of interest. Traditionally, AC methods have been used to tune the parameters of specific algorithms~\cite{Birattari09tuning}, however, more recently, they have been used to automatically design algorithms from flexible frameworks~\cite{LopStu2012tec,KhuXuHooLey16:aij,MasLopDubStu2014cor,AziDoeDre2021} and to build algorithm portfolios~\cite{XuHooLey2010aaai}. For more background on AC, we refer the interested reader to two recent surveys~\cite{reviewer1,reviewer2}. 

As mentioned above, we do not wish to re-build nor replace NGOpt but, instead, iteratively improve it. To do so, we first answer the question of whether the hand-crafted NGOpt may be further improved by replacing some of its components by algorithm configurations obtained from the application of an AC method, in particular, irace. When applied to a complex algorithm selection wizard such as NGOpt and a diverse black-box optimization scenario, such as the one tackled by NGOpt, the correct setup of an AC method is crucial. We discuss in detail our approach here, which we believe is applicable to other similar scenarios. Our results show that our proposed approach clearly improves the
hand-crafted NGOpt, even on benchmark suites not considered in the tuning process by irace.

This paper is structured as follows. Section~\ref{sec:preliminaries} gives an introduction to the concept of algorithm selection wizard and describes, in particular, the one used in our experiments, NGOpt. Section~\ref{sec:experimental_setup} presents the experimental setup and the benchmark suites used for tuning the target algorithm. In Section~\ref{sec:metacma} we present the integration of the tuned algorithms into the new algorithm selection wizard. The evaluation results are presented in Section~\ref{sec:experimental_results}. Section~\ref{sec:conclusion} concludes the paper.

\section{Preliminaries}\label{sec:preliminaries}
\textbf{Algorithm Selection Wizards.} 
Modern algorithm selection wizards make use of a number of different criteria to chose which algorithm(s) are executed on a given problem instance, and for how long. Algorithm selection wizards take into account a priori information about the problem and the resources that are available to solve it. Using this information, the algorithm selection wizards recommend one or several algorithms to be run, along with a protocol to assign computational budget to each of these. 

NGOpt is built atop of several dozens of state-of-the-art algorithms that have been added to Nevergrad over the last years. However, NGOpt does not only select which algorithms to execute on which problem instances, but it also combines algorithms in several ways, e.g., by enriching classic approaches with local surrogate models. 
Since all algorithms submitted to Nevergrad are periodically run on all fitting benchmark problems available within the environment, a very large amount of performance data is available. 
In light of this rich data set, and in light of the tremendous performance gains obtained through automated algorithm designs~\cite{KhuXuHooLey16:aij,BezLopStu2019ec,PagStu2019:ejor} %
and selection rules~\cite{KerHooNeuTra2019}, it is therefore surprising that both the decision rules of NGOpt and the configuration of the algorithms available in Nevergrad are hand-picked.

\textbf{CMA.} 
Among the algorithms that compose NGOpt, a key component is CMA, an instance of the family of covariance matrix adaptation evolution strategies (CMA-ES~\cite{HanOst2001ec}). 
In Nevergrad's CMA implementation, the following parameters are explicitly exposed, making them straightforward candidates for the configuration task.
The \texttt{scale} parameter controls the scale of the search of the algorithm in the search domain: a small value leads to start close to the center. %
The \texttt{elitist} parameter is a Boolean variable that can have values `True' or `False' and it represents a switch for elitist mode, i.e., preserving the best individual even if it includes a parent.
The \texttt{diagonal} parameter is another Boolean setting that controls the use of diagonal CMA~\cite{RosHan2008diagcma}.
Finally, the population size is another crucial parameter that is usually set according to the dimension of the problem. Instead of setting its value directly, we decided to create a higher-level integer parameter 
  \texttt{popsize\_factor} and let the population size be $\lfloor 4 + \texttt{popsize\_factor} \cdot \log d\rfloor$, where $d$ is the problem dimension.
The parameter search space of CMA is shown in Table~\ref{tab:cmaalg}.

\textbf{Automated Algorithm Configuration.}
 %
Given a description of the parameter space of a target algorithm, a class of
problems of interest, and a computational budget (typically measured in target
algorithm runs), the goal of an AC method is to find a parameter configuration
of the target algorithm that is expected to perform well on the problem class
of interest. Due to the stochasticity of randomized algorithms and the fact that only a limited sample of (\emph{training}) problem instances can be evaluated in practice, specialized methods for AC have been developed in recent years~\cite{HuaLiYao2020algconf,reviewer1,reviewer2} that try to avoid inherent human biases and limitations in manual parameter tuning. 
We have selected here the \emph{elitist iterated racing} algorithm
implemented by the irace package~\citep{LopDubPerStuBir2016irace}, since it has shown good performance for configuring black-box optimization algorithms in similar scenarios~\cite{LopLiaStu2012ppsn,LiaMolStu2015,LiaStuMonDor2014,LiaMonStu13:soco}.
\newcommand{\FirstTest}{\emph{FirstTest}\xspace}
\newcommand{\EachTest}{\emph{EachTest}\xspace}

\section{Experimental Setup}\label{sec:experimental_setup}
To show the influence of the parameter tuning of the CMA included in NGOpt~\cite{nevergrad}, we have performed several parameter tuning experiments using irace. The setup and results of this process are described in this section.

\subsection{Setup of Irace for Tuning CMA}

A training \emph{instance} is defined by a benchmark function, a dimension, a
rotation and the budget available to CMA. We also define ``blocks'' of instances: all instances within a block are equal except for the
benchmark function and there are as many instances within a block as benchmark
functions.  %
We setup irace so that, within each race, the first elimination test
(\FirstTest) happens after seeing 5 blocks and subsequent elimination tests
(\EachTest) happen after each block. Moreover, configurations are evaluated by irace on
blocks in order of increasing budget first and increasing dimension second,
such that we can quickly discard poor-performing configurations on small
budgets and only good configurations are evaluated on large ones~\cite{StyHoo2013gecco}.
The performance criterion optimized by irace is the objective value of the point recommended by CMA after it has exhausted its budget.
Since Nevergrad validates performance according to the mean loss (as explained later), the elimination test used by irace is set to t-test.
Finally, we set a maximum of 10\,000 individual runs %
of CMA as the termination
criterion of each irace run. By parallelizing each irace run across 4 CPUs, the
runtime of a single run of irace was around 8 hours. %

\subsection{Benchmark Suites Used for Tuning}
Nevergrad contains a very large number of benchmark \emph{suites}. Each suite is a collection of benchmark problems that share some common properties such as similar domains or constraints, similar computational budgets, etc. A large number of different academic and real-world problems are available. %
We selected five suites (\YABBOB, \YASMALLBBOB, \YATUNINGBBOB, \YAPARABBOB, and \YABOUNDEDBBOB) as training instances for irace, all of them derived from the BBOB benchmark suite~\cite{HanAugFin2009bbob_setup} of the COCO benchmarking environment~\cite{HanAugMer2020coco}. See Table~\ref{tab:trainbench} for their various characteristics.
We ran irace once for each benchmark suite separately, by setting up their
problem instances as described above. All runs of irace used the parameter
space of CMA shown in Table~\ref{tab:cmaalg}.
For each benchmark suite, we selected one CMA configuration from the ones returned by irace by doing 10 validation runs using the best configurations obtained by irace and the default CMA configuration. Each run consisted of running each CMA configuration on the whole instance space. For each instance we declared a winner configuration that yielded the smallest result. Using majority vote then we determined a winner from the 10 validation runs.  The optimized configurations selected are named  CMAstd for \YABBOB, CMAsmall for \YASMALLBBOB, CMAtuning  for \YATUNINGBBOB, CMApara for \YAPARABBOB, and CMAbounded for \YABOUNDEDBBOB (Table~\ref{tab:trainbench}). Their parameter values are shown in Table~\ref{tab:iraceConf}.

\begin{table}[t]\scriptsize
    \caption{\label{tab:trainbench}Artificial benchmarks used as training instances by irace. The third column gives the name of the CMA variant obtained by irace on that particular benchmark (for the parameter values, see Table~\ref{tab:iraceConf}). The detailed code can be found in \cite{nevergrad}.  %
  \label{tab:testbench}}
  
     \centering%
\begin{tabular}{ccc}
     \toprule
     \bf Benchmark name & \bf Context                                       & \bf Optimized  CMA \\
     \midrule
     \YABBOB            & dimension $\in [2,50]$, budget $\in [50, 12800]$  & CMAstd             \\
     \YASMALLBBOB       & Budget $<50$                                      & CMAsmall           \\
     \YATUNINGBBOB      & Budget $<50$ and dim $\leq15$                     & CMAtuning          \\
     \YAPARABBOB        & Num-workers=100                                   & CMApara            \\
     \YABOUNDEDBBOB     & Box-constrained, budget $\leq 300$, dim $\leq 40$ & CMAbounded         \\
     \bottomrule
   \end{tabular}
   \end{table}

\begin{table}

  \caption{Artificial and real-world benchmarks used for testing. The detailed code can be found in \cite{nevergrad}.}

\begin{tabular}{lcl}
  \toprule
  \bf Type & \bf Name & \bf Context                                        \\
  \midrule
 Artificial & \YABIGBBOB            & {Budget 40000 to 320000 }                                              \\
  &\YABOXBBOB            & {Box constrained }                                                     \\
  &\YAHDBBOB             & {Dimension 100 to 3000}                                                \\
  &\YAWIDEBBOB           & {Different settings (multi-objective, noisy, discrete\ldots) } \\
    &Deceptive             & {Hard benchmark, far from \YABBOB}                                      \\
  \midrule
  Real-world&SeqMLTuning           & {Hyperparameter tuning for SciKit models}                              \\
  &SimpleTSP             & {Traveling Salesman Problem, black-box}                                \\
  &ComplexTSP            & {Traveling Salesman Problem, black-box, nonlinear terms}               \\
  &UnitCommitment        & {Unit Commitment for power systems}                                    \\
  &Photonics             & {Simulators of nano-scale structural engineering}                      \\
  &GP                    & {Gym environments used in \cite{VidLeiTey2022eurogp}.}                 \\
  &007                   & Game simulator for the 007 game \\
  \bottomrule
\end{tabular}
\end{table}
\begin{table}\caption{  \label{tab:cmaalg}Search space for the tuning of CMA by irace.}

  \centering
  \begin{tabular}{r@{\hskip 1em}c@{\hskip 1em}c}
    \toprule
    \bf Parameter      & \bf Default value & \bf Domain         \\
    \midrule
    scale          & 1             & $(0.1, 10) \subset \mathbb{R}$     \\
    popsize-factor & 3             & $[1,9] \subset \mathbb{N}$         \\
    elitist        & False         & \{True, False\} \\
    diagonal       & False         & \{True, False\} \\
    \bottomrule
  \end{tabular}
\end{table}

\begin{table}[!t]
  \caption{\label{tab:iraceConf}Best parameters found by irace for each benchmark suite.}
  \centering\scriptsize%
  \resizebox{\linewidth}{!}{%
	\begin{tabular}{rccccc}
           \toprule
          & \bf CMAstd & \bf CMAsmall & \bf CMAtuning & \bf CMApara & \bf CMAbounded\\
		\bf Parameter & \bf (Yabbob) & \bf(Yasmallbbob) & \bf(Yatuningbob) & \bf(Yaparabbob) & \bf(Yaboundedbbob) \\
		 \midrule
		scale & 0.3607 & 0.4151 & 0.4847 & 0.8905 & 1.5884 \\
		popsize-factor & 3 & 9 & 1 & 8 & 1 \\
		elitist & False & False & True & True & True\\
		diagonal & False & False & False & True & True \\
		\bottomrule
	\end{tabular}}
\end{table}

\section{MetaCMA and its Integration in Nevergard}\label{sec:metacma}
The CMA configurations found by irace were further combined into a new algorithm selection wizard.
For this purpose, we propose MetaCMA (Algorithm~\ref{alg:metacma2}), a deterministic ensemble for noiseless single-objective continuous optimization that, based on deterministic rules, switches between different tuned CMA configurations. %
For problems where all variables have bounded domain, MetaCMA selects CMAbounded. Otherwise, for low budget (function evaluations) available, the MetaCMA model selects either the CMAtuning configuration (for low dimension) or the CMAsmall configuration (otherwise). If number of workers is more than 20, then the MetaCMA uses the CMApara. If neither of the above-mentioned rules is met, the MetaCMA will switch to CMAstd, which was tuned on \YABBOB.

Given that NGOpt considers many more cases than MetaCMA, we have to integrate it inside NGOpt so that we have both the improved performance of our MetaCMA and the generality of NGOpt.
NGTuned corresponds to NGOpt with all instances of CMA in NGOpt replaced by MetaCMA.

\begin{algorithm}[!tbp]
\caption{Pseudocode for MetaCMA.}
\label{alg:metacma2}
\begin{algorithmic}[1]
\Procedure{MetaCMA}{$budget, dimension, num\_workers, fully\_bounded$}       
    \If{$fully\_bounded$} \Return CMAbounded\Comment{All variables have bounded domain.}
    \EndIf
    \If{$budget < 50$}
    \IfThenElse{$dimension \leq 15$}{\Return  CMAtuning}{\Return CMAsmall}
    \EndIf
    \IfThenElse{$num\_workers > 20$}{\Return CMApara}{\Return CMAstd}
\EndProcedure
\end{algorithmic}
\end{algorithm}

\section{Experimental Results}\label{sec:experimental_results}
To evaluate the performance of the MetaCMA model and its integration in NGOpt (NGTuned), we compared it with the previous variant of NGOpt and other state-of-the-art algorithms in three different scenarios. The first scenario consists of the benchmark suites that were used by irace during tuning. All  algorithms have been rerun independently of the tuning process.
The second scenario involves artificial benchmark suites whose problem instances were not used in the tuning process.
The third scenario involves real-world problems not presented in the tuning process of irace. Details about these benchmark suites are given in Table~\ref{tab:trainbench}. %

The selection of algorithms involved in the evaluation for each benchmark suite differ because each suite has its own set of state-of-the-art algorithms.

\begin{figure}[!tp]
    \centering
    \includegraphics[width=.49\textwidth]{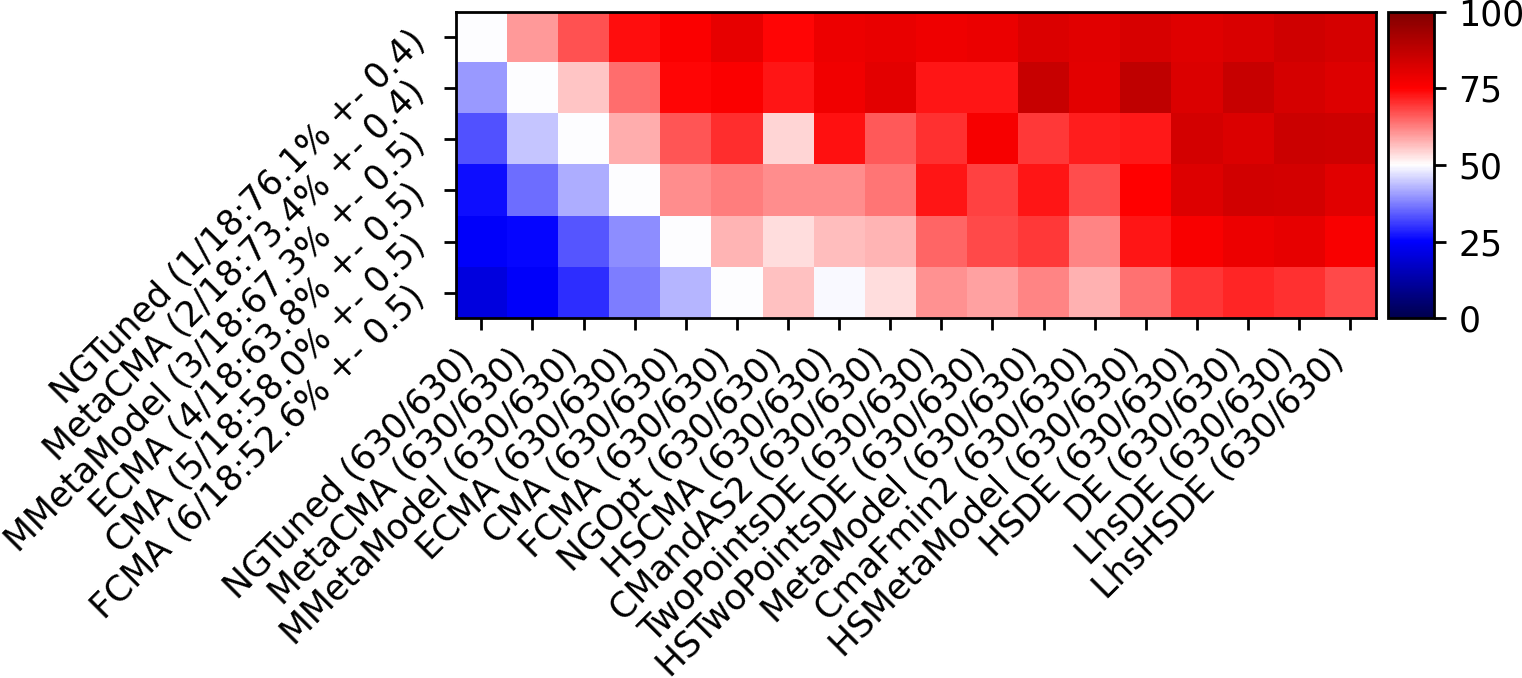}
    \includegraphics[width=.49\textwidth]{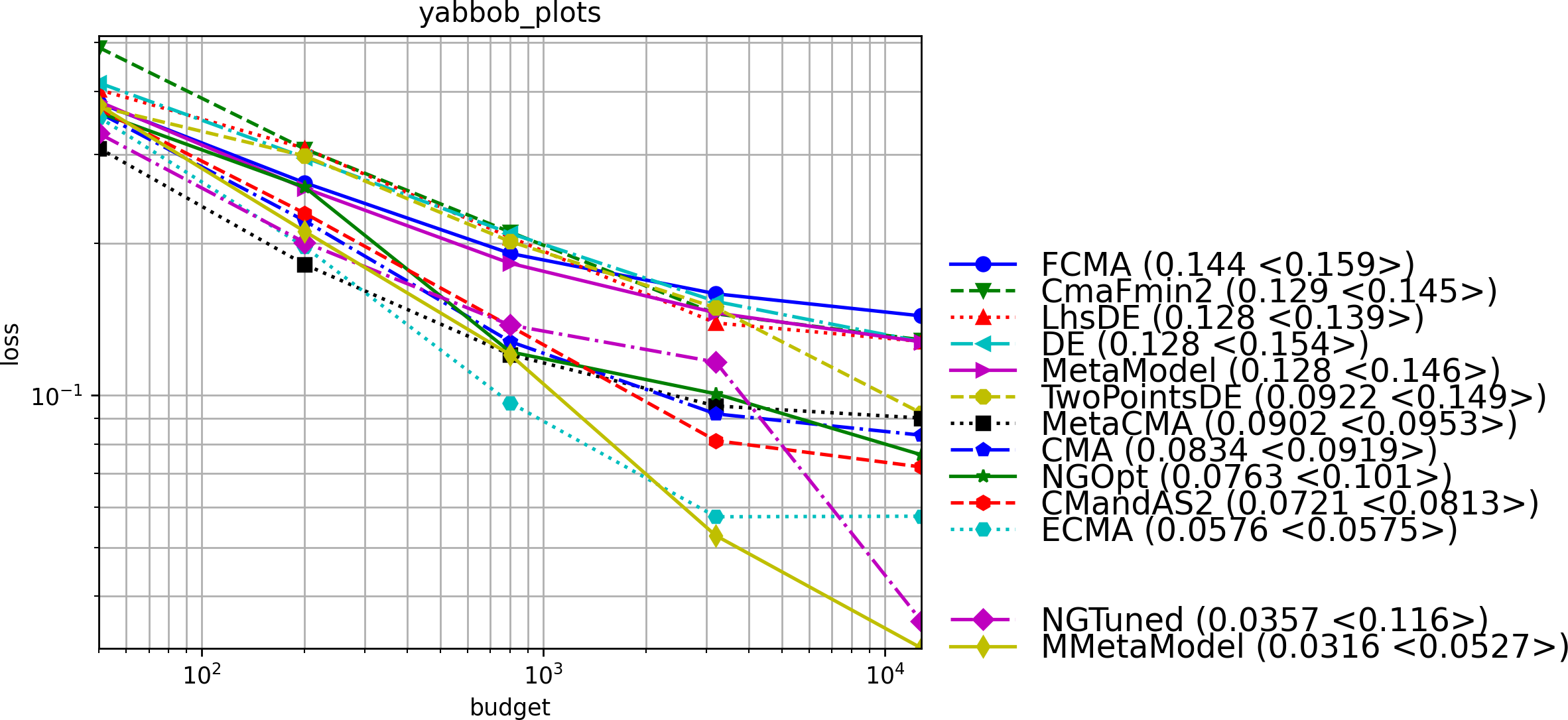}
    \caption{Results on \YABBOB: suite used in the tuning, moderate dimension and budget.}
    \label{yabbob}
\end{figure}
\begin{figure}[!tp]
\centering
\includegraphics[height=3cm,width=.49\textwidth]{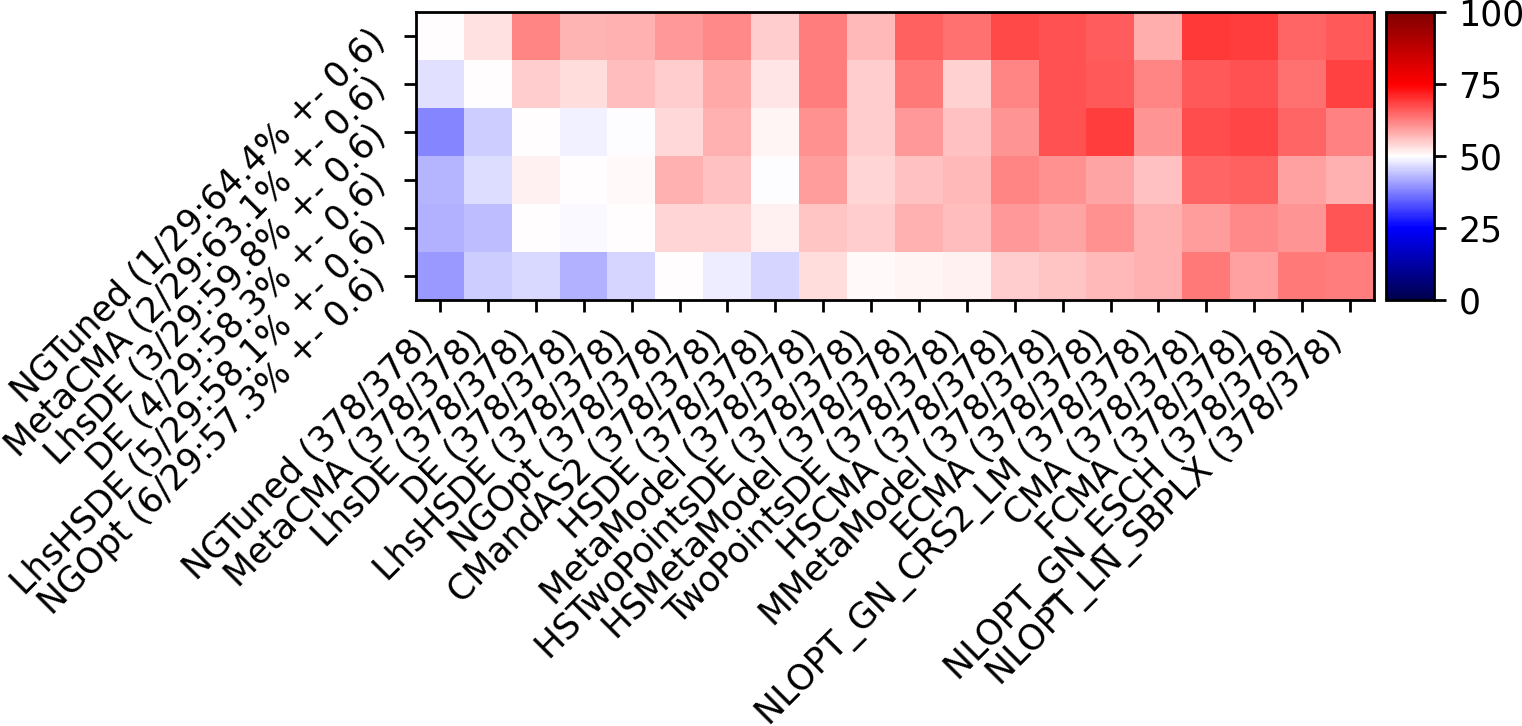}
\includegraphics[height=3cm,width=.49\textwidth]{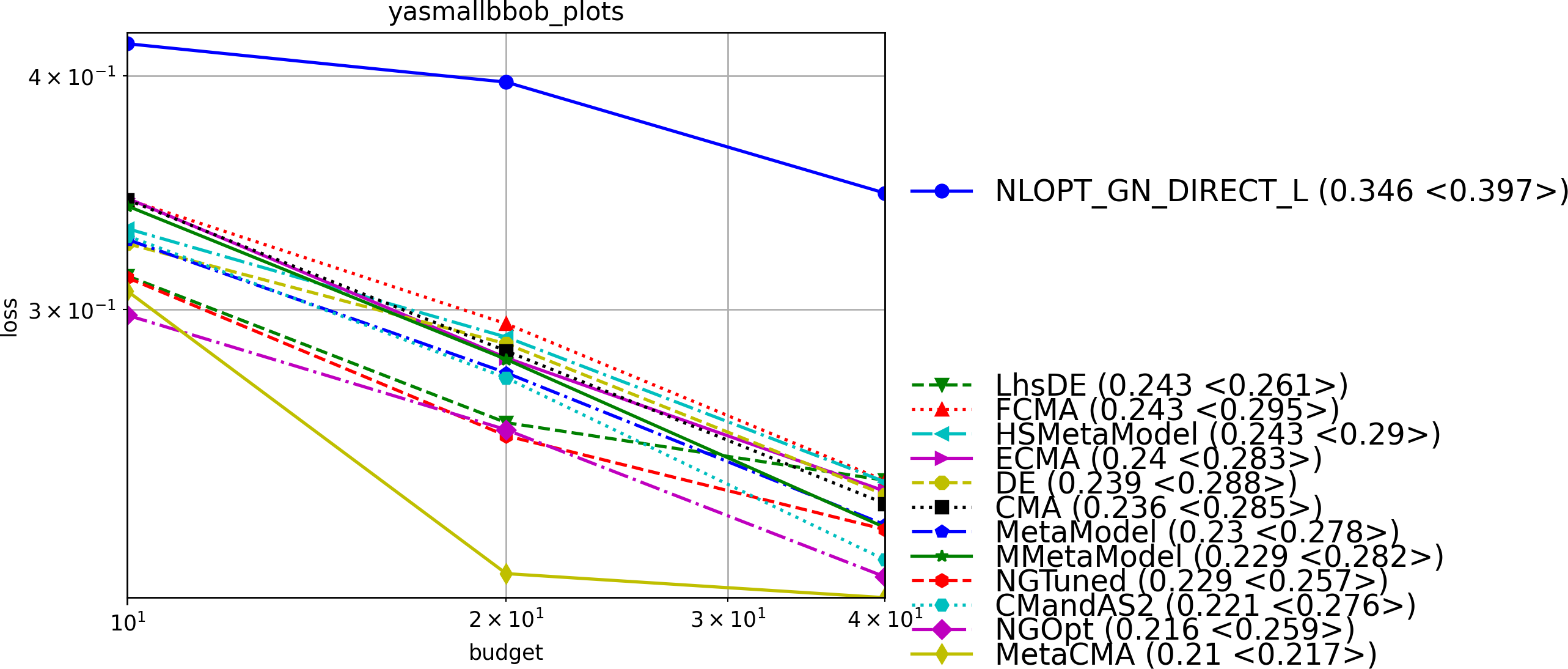}
\caption{Results on \YASMALLBBOB: benchmark suite used in the tuning, low budget.}
\label{yasmallbbob}
\end{figure}
\begin{figure}[!tp]
\centering
\includegraphics[height=3cm,width=.49\textwidth]{{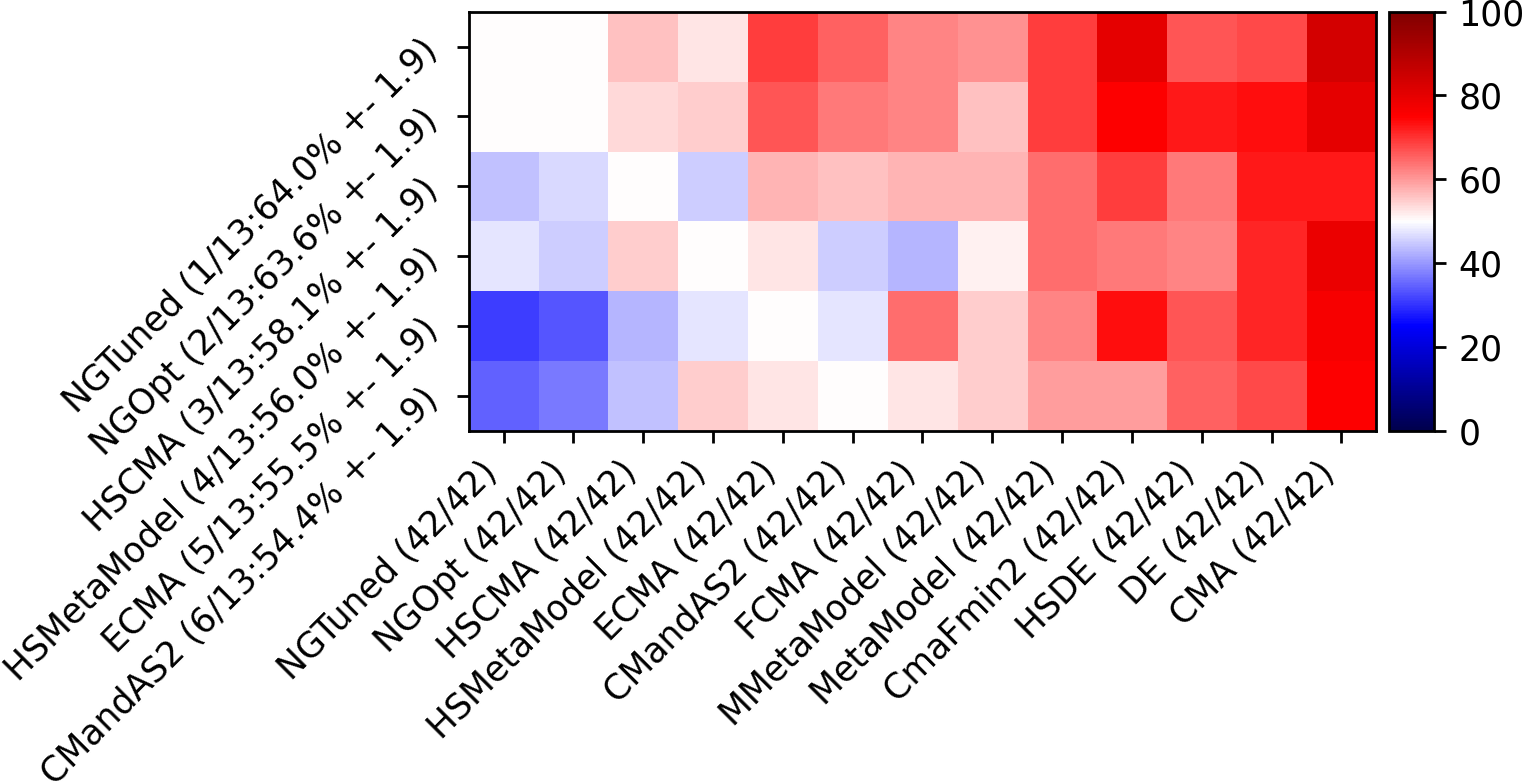}}
\includegraphics[height=3cm,width=.49\textwidth]{{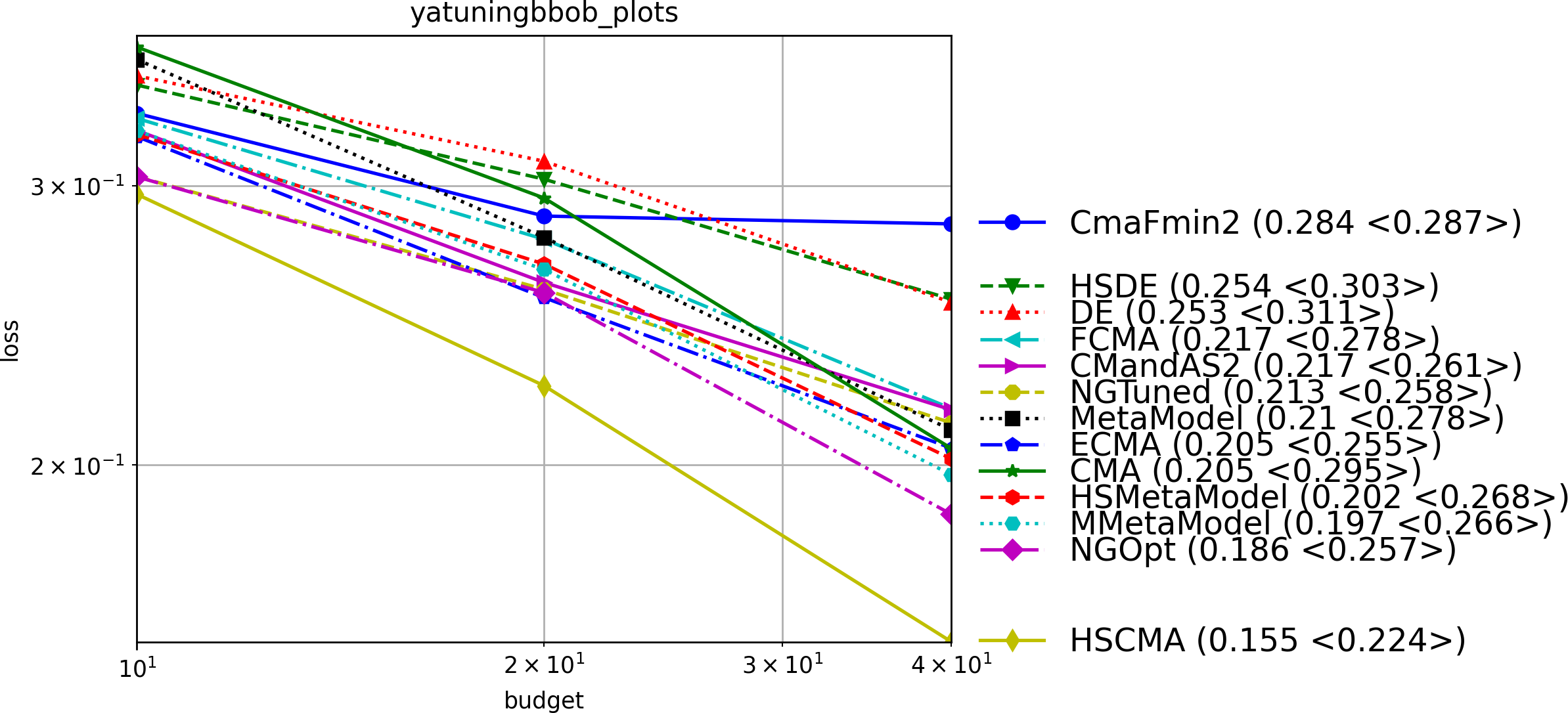}}
\caption{Results on \YATUNINGBBOB: benchmark suite used in the tuning, counterpart of \YABBOB with low budget and low dimension.}
\label{yatuningbbob}
\end{figure}
\begin{figure}[tp]
    \centering
    \includegraphics[height=3cm,width=.45\textwidth]{{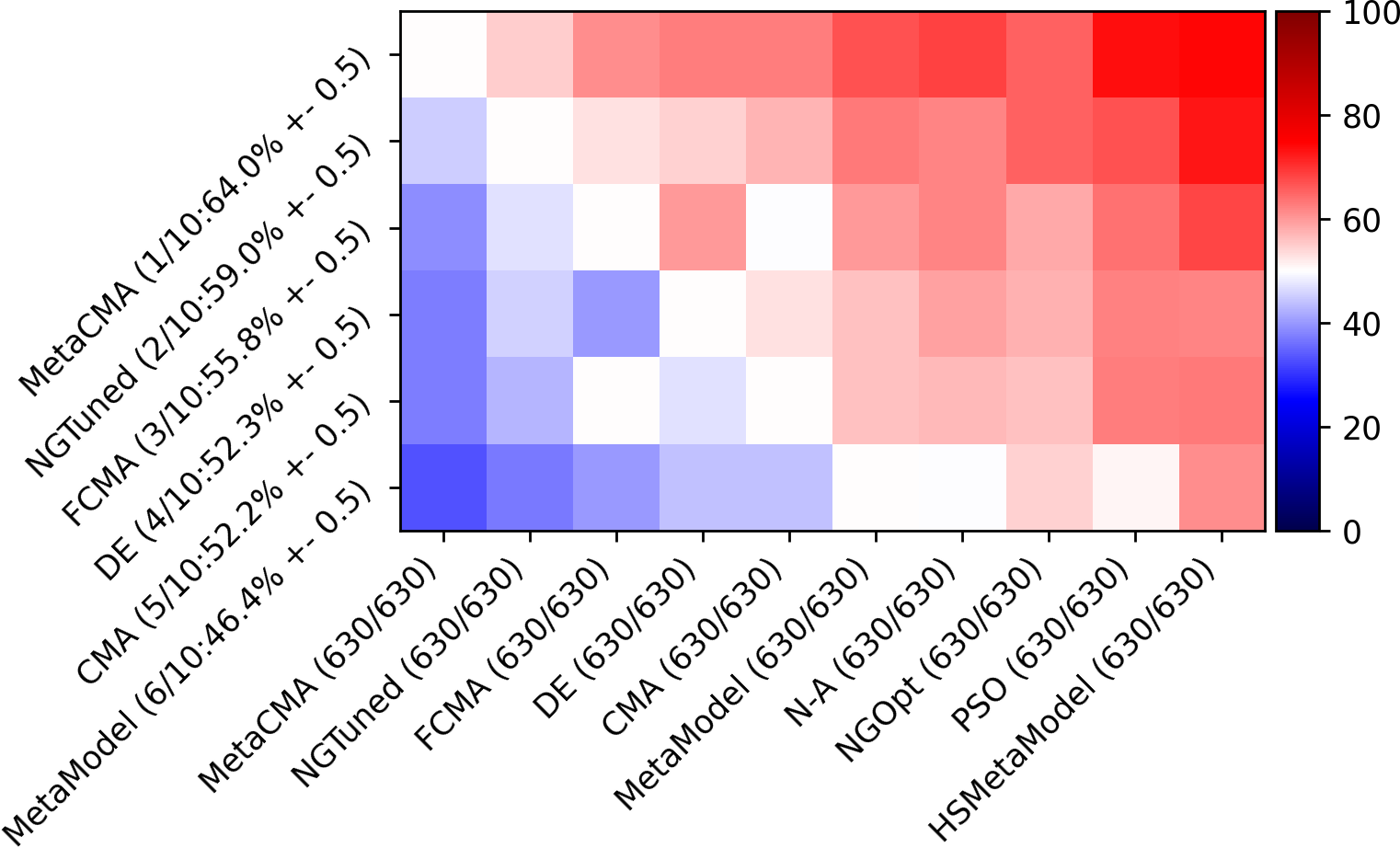}}
    \includegraphics[height=3cm,width=.49\textwidth]{{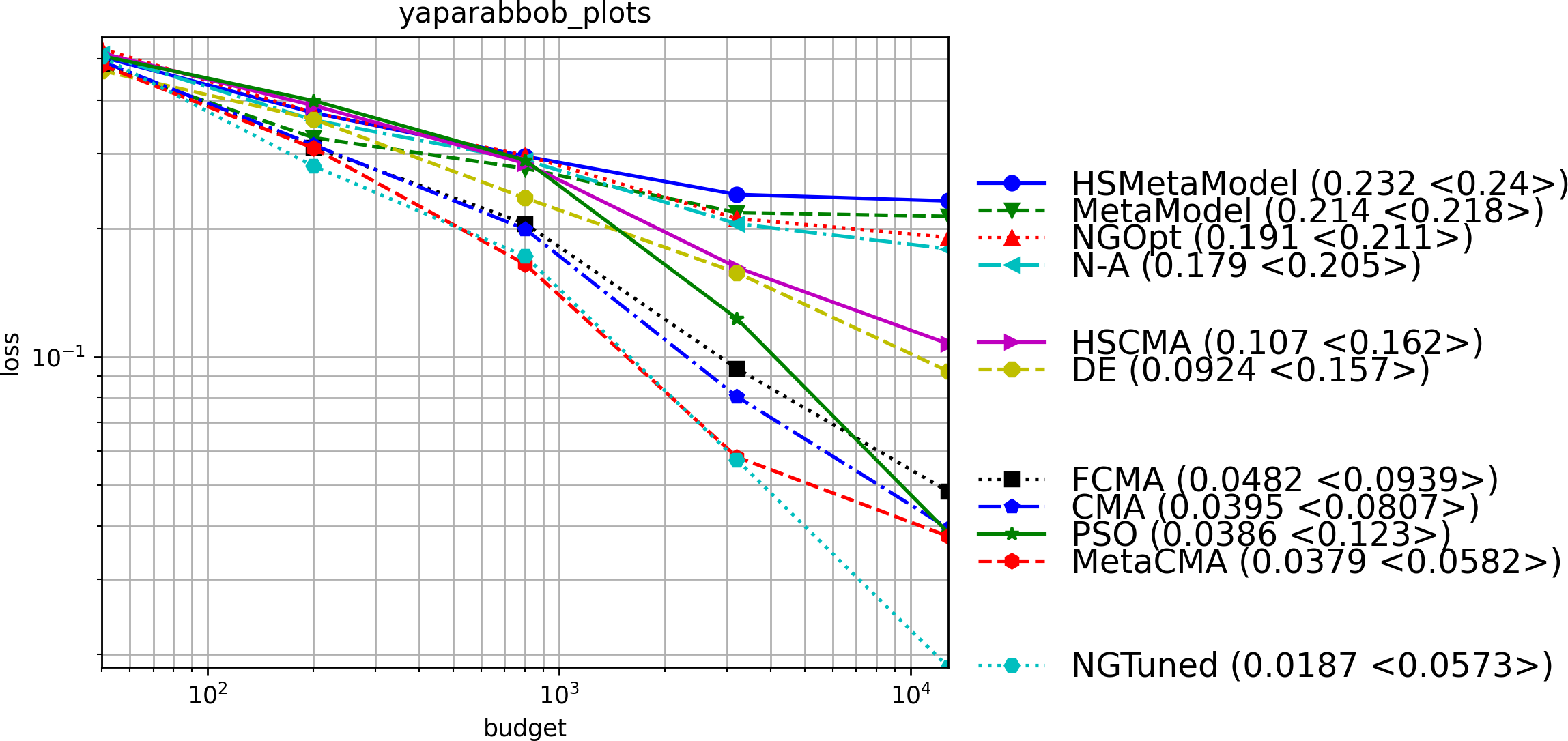}}
    \caption{Results on \YAPARABBOB: benchmark suite used in the tuning, parallel case.}
    \label{yaparabbob}
\end{figure}
\begin{figure}[tp]
    \centering
    \includegraphics[height=3cm,width=.45\textwidth]{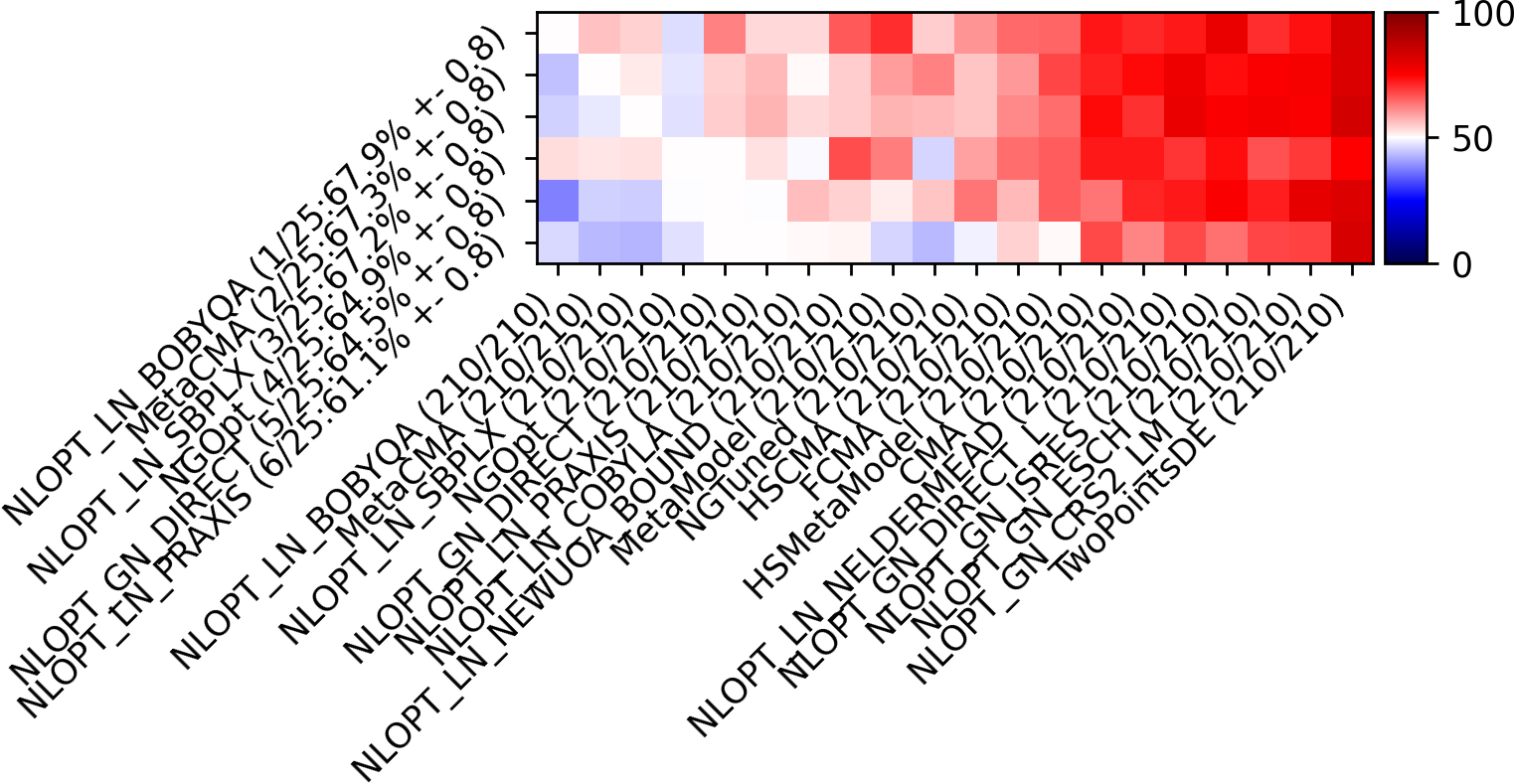}
    \includegraphics[height=3cm,width=.49\textwidth]{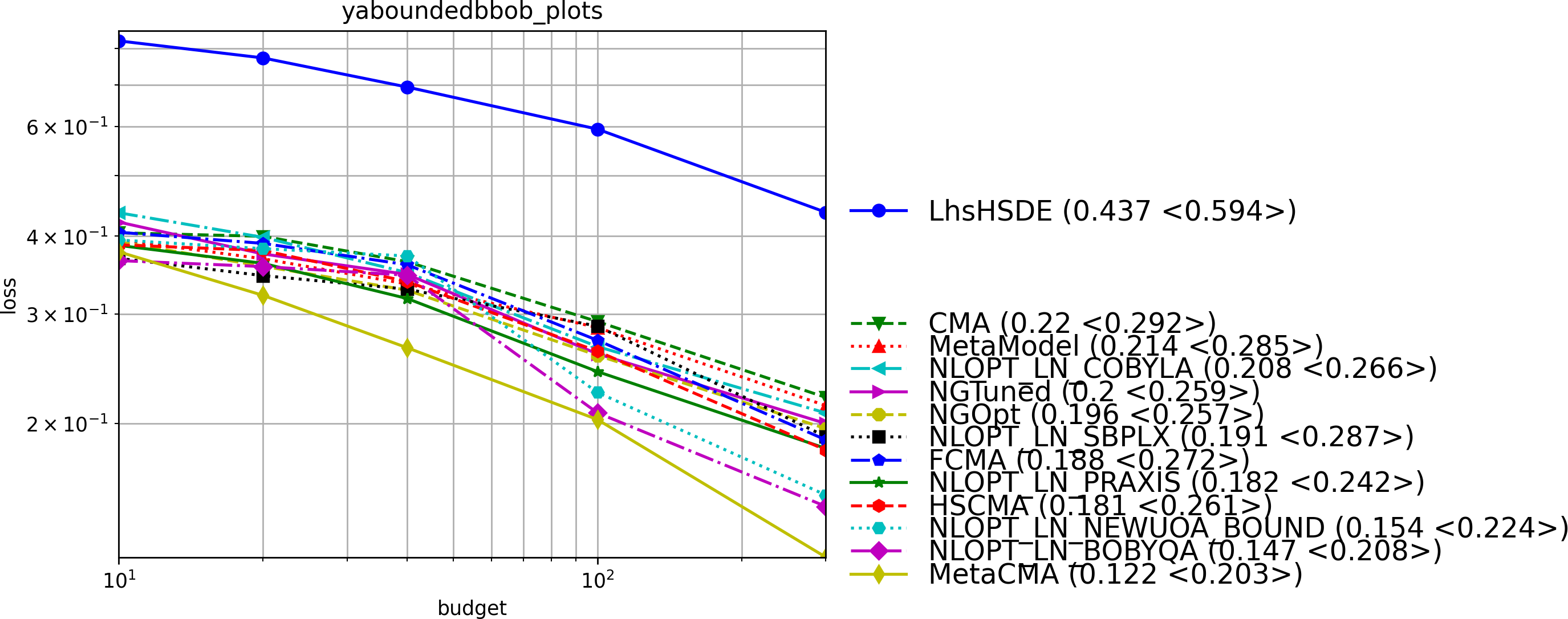}
    \caption{\YABOUNDEDBBOB: suite used in the tuning, bounded domain.}
    \label{yaboundedbbob}
  \end{figure}

\subsection{Evaluation on Artificial Benchmark Suites Presented in the Tuning Process of Irace}
Figures~\ref{yabbob}, \ref{yasmallbbob}, \ref{yatuningbbob}, \ref{yaparabbob}, and \ref{yaboundedbbob}  present results on benchmark suites used for running irace. 
On the left side of each figure is a robustness heatmap, where the rows and columns represent different algorithms, and the values of the heatmap reflect the number of times the algorithm in each row outperforms the algorithm in each column. We use mean quantiles over multiple settings (compared to other optimization methods) in an anytime manner, meaning that all budgets are taken into account, not only the maximum one. This is robust to arbitrary scalings of objective functions. The algorithms' global score, defined as their mean  scores (the number of times it performs better than the other algorithms), is used for ranking the rows and columns. The six best-performing algorithms are listed as rows and sorted by their mean score, while the columns contain all of the algorithms involved in the algorithm portfolio.
The labels in the columns contain the mean score of each algorithm.
For instance, in Figure~\ref{yabbob}, the label for NGTuned is ``1/18:76.1\% +- 0.4", meaning that it was ranked first, out of 18 evaluated algorithms, and its mean score is 76.1\%, which is the mean number of times it performed better than the rest of the algorithms, on all problems and all budgets in the benchmark suite.
On the right side of each figure, we present the algorithms' performance with different budgets. The x-axis represents the budget, while the y-axis represents the rescaled losses (normalized objective value, averaged over all settings), for a given budget, and linearly normalized in the range $[0, 1]$.
 Apart from the algorithm name, the label of each algorithm in the legend contains two values in brackets. The first one denotes the mean loss achieved by the algorithm for the maximum budget (after the losses for each problem have been normalized in the range $[0, 1]$), while the second one denotes the mean loss for the second largest budget.
Figure~\ref{yabbob} features the results on the \YABBOB benchmark suite. The heatmap (left) shows that NGTuned outperforms the other algorithms, followed by MetaCMA, while NGOpt is in the 7th position.%
Figure~\ref{yasmallbbob} features the results on the \YASMALLBBOB benchmark suite. The best results are obtained by NGTuned, followed by MetaCMA, while NgOPt is ranked in the 6th position. In addition, with regard to different budgets, MetaCMA outperforms NGOpt by obtaining lower losses.
In Figure~\ref{yatuningbbob}, the results from the evaluation on \YATUNINGBBOB benchmark suite are presented. It follows the trend of NGTuned being the best performing algorithm. 
Comparing NGTuned and NGOpt according to the convergence it seems that till 2$\times 10^1$, both achieved similar loss, but later on for budgets above 2$\times 10^1$, NGOpt can provide better losses. The comparison between the left and the right plot (keeping the rescaling in mind) suggests that NGTuned has a better rank than NGOpt, but that the rescaled loss is better for NGOpt.
The results for \YAPARABBOB benchmark suite are presented in Figure~\ref{yaparabbob}. In terms of mean scores obtained across all budgets, MetaCMA is the best algorithm, followed by NGTuned. In terms of the loss achieved for different budgets, NGTuned and MetaCMA provide similar results, with the biggest difference being in the largest budget, where NGTuned achieves a lower loss. Due to rescalings and the difference criteria (ranks on the left, normalized loss on the right) the visual comparison between algorithms can differ.
Both outperform NGOpt. 
On the \YABOUNDEDBBOB benchmark suite (Figure~\ref{yaboundedbbob}), NGOpt achieves better performance than NGTuned. Here, several algorithms from \cite{nlopt} have been tested: the low budget makes them relevant. The best results are achieved by NLOPT\_LN\_BOBYQA~\cite{bobyqa}, %
closely followed by MetaCMA, with a difference of 0.4\% in the average quantiles referring to the number of times they outperform other methods. %
From the line plot on the right of the figure, we can see that MetaCMA consistently provides the lowest normalized loss, for different budgets. %

\afterpage{\clearpage
\begin{figure}[!t]\centering\includegraphics[height=3cm,width=.49\textwidth]{{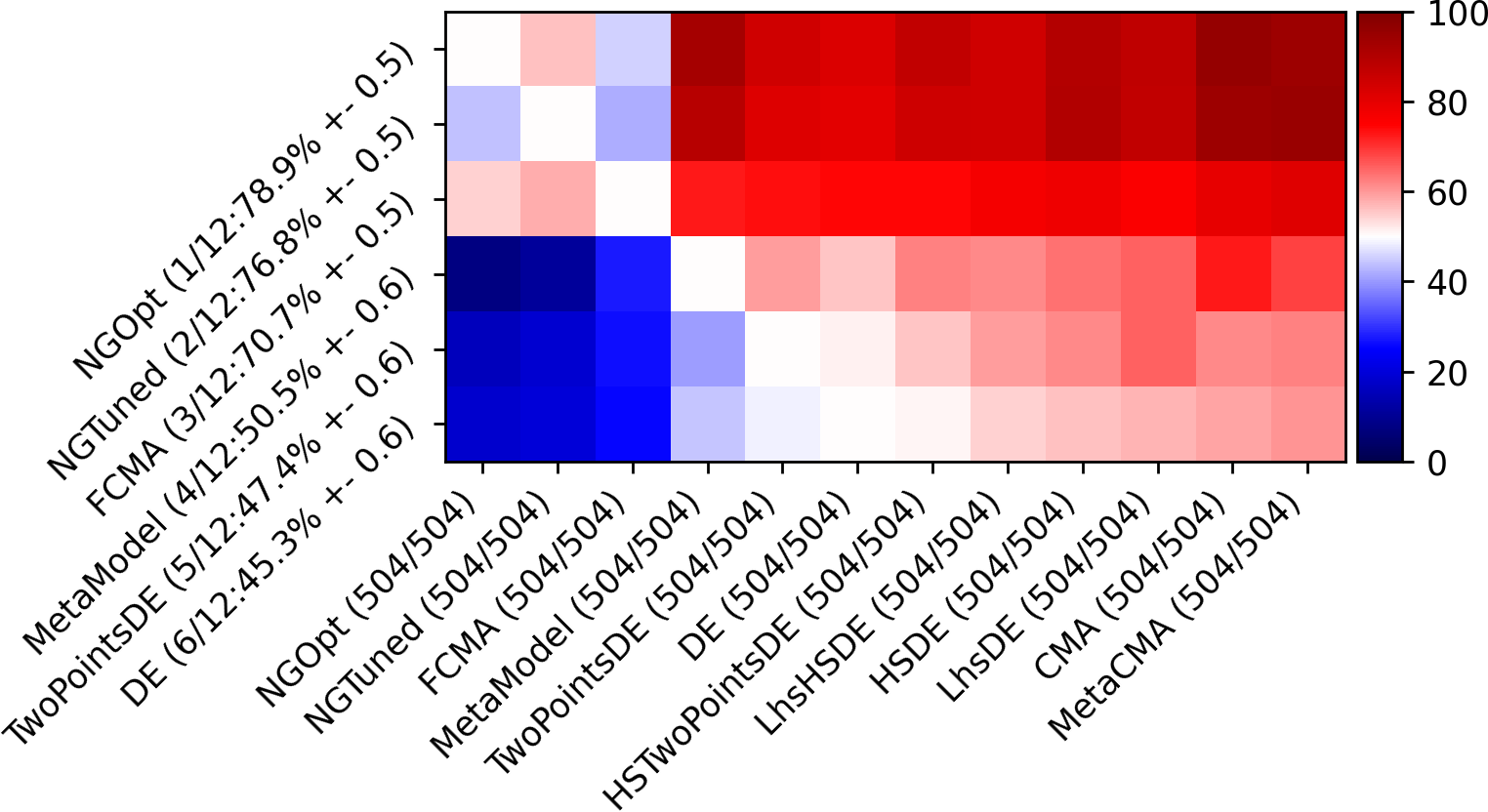}}
\includegraphics[height=3cm,width=.49\textwidth]{{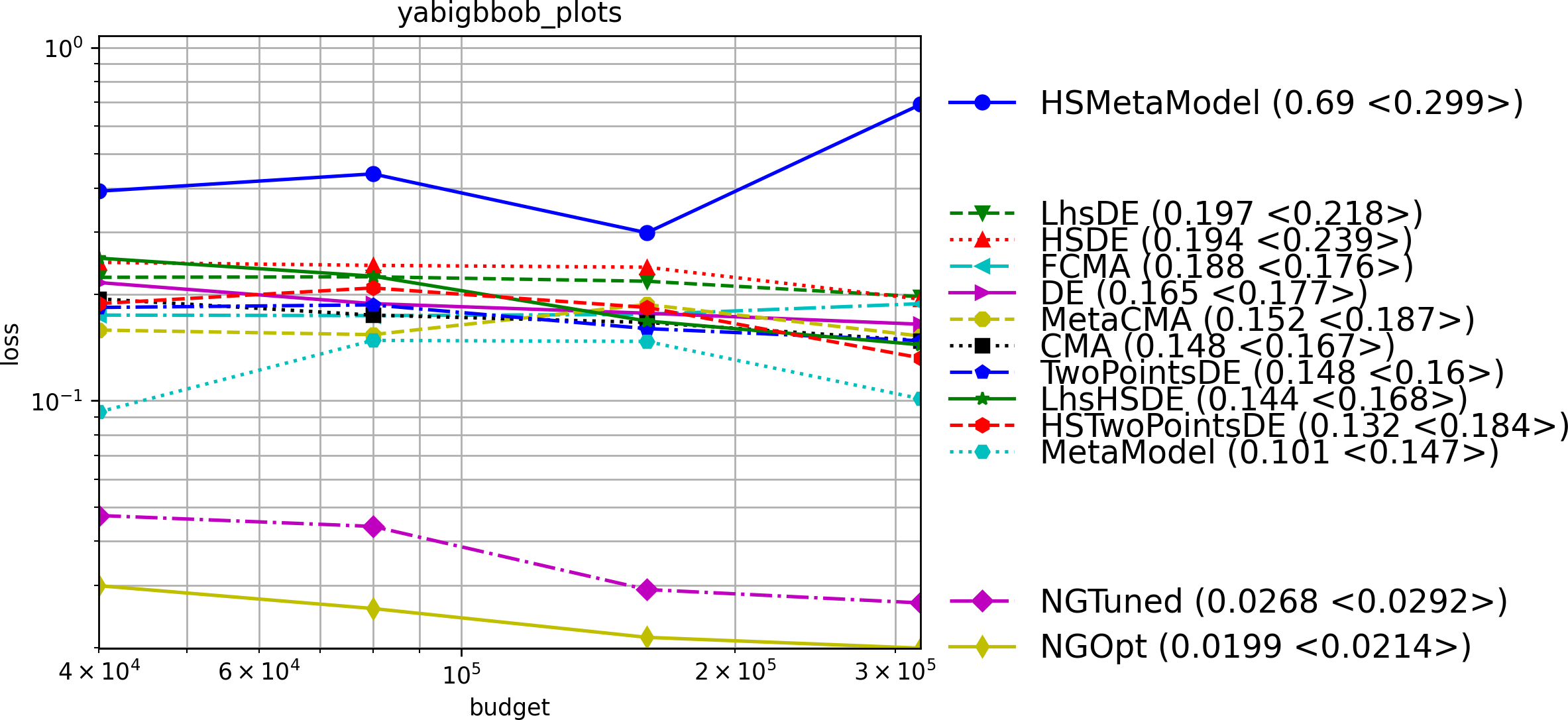}}
\caption{Results on \YABIGBBOB: longer budgets than for BBOB. Not used in the irace trainings, but same functions as \YABBOB.}\label{yabigbbob}
\end{figure}
\begin{figure}[!t]\centering\includegraphics[height=3cm,width=.49\textwidth]{{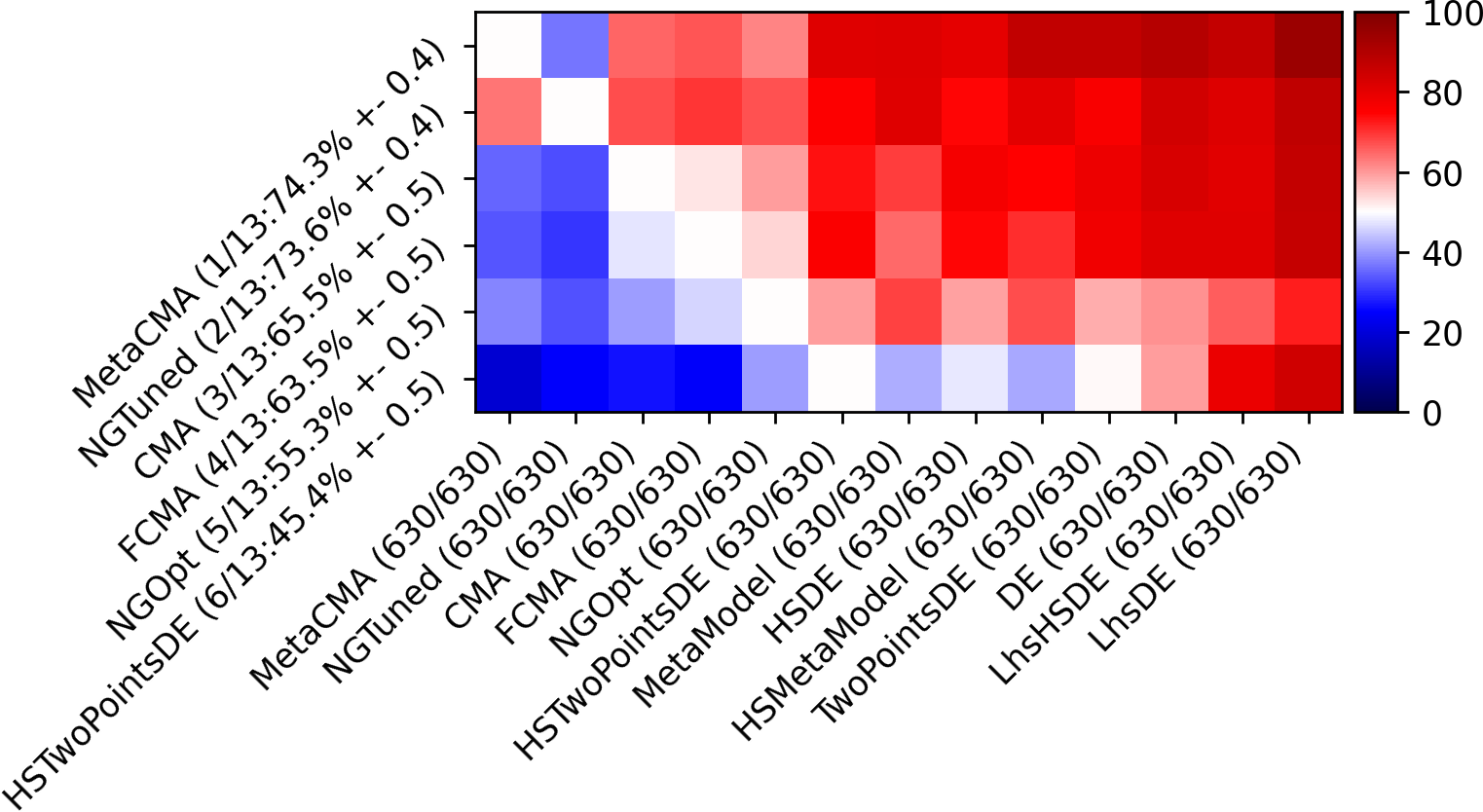}}
\includegraphics[height=3cm,width=.49\textwidth]{{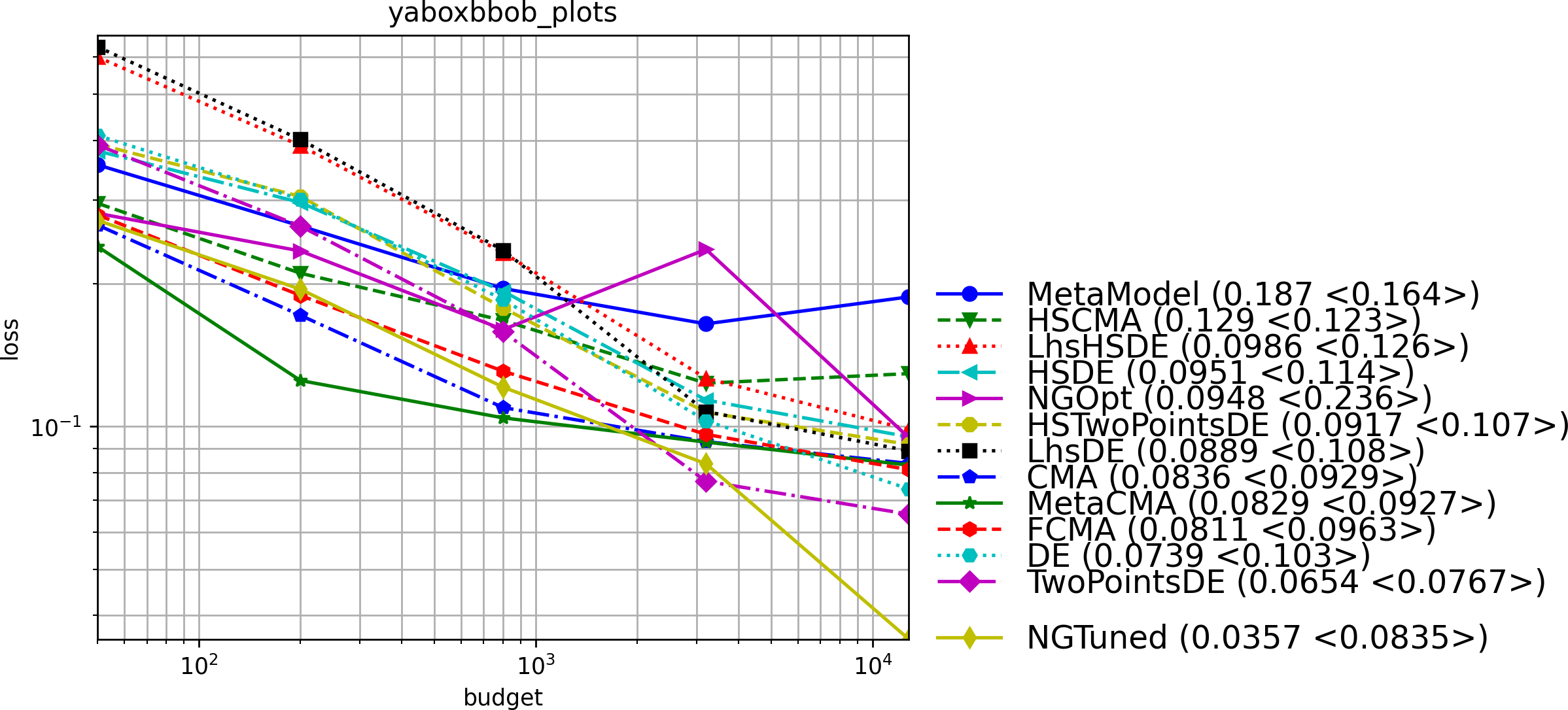}}\caption{Results on \YABOXBBOB: box-constrained and low budget: not used in the irace training but has similarities.}\label{yaboxbbob}\end{figure}
\begin{figure}[!t]\centering\includegraphics[height=3cm,width=.49\textwidth]{{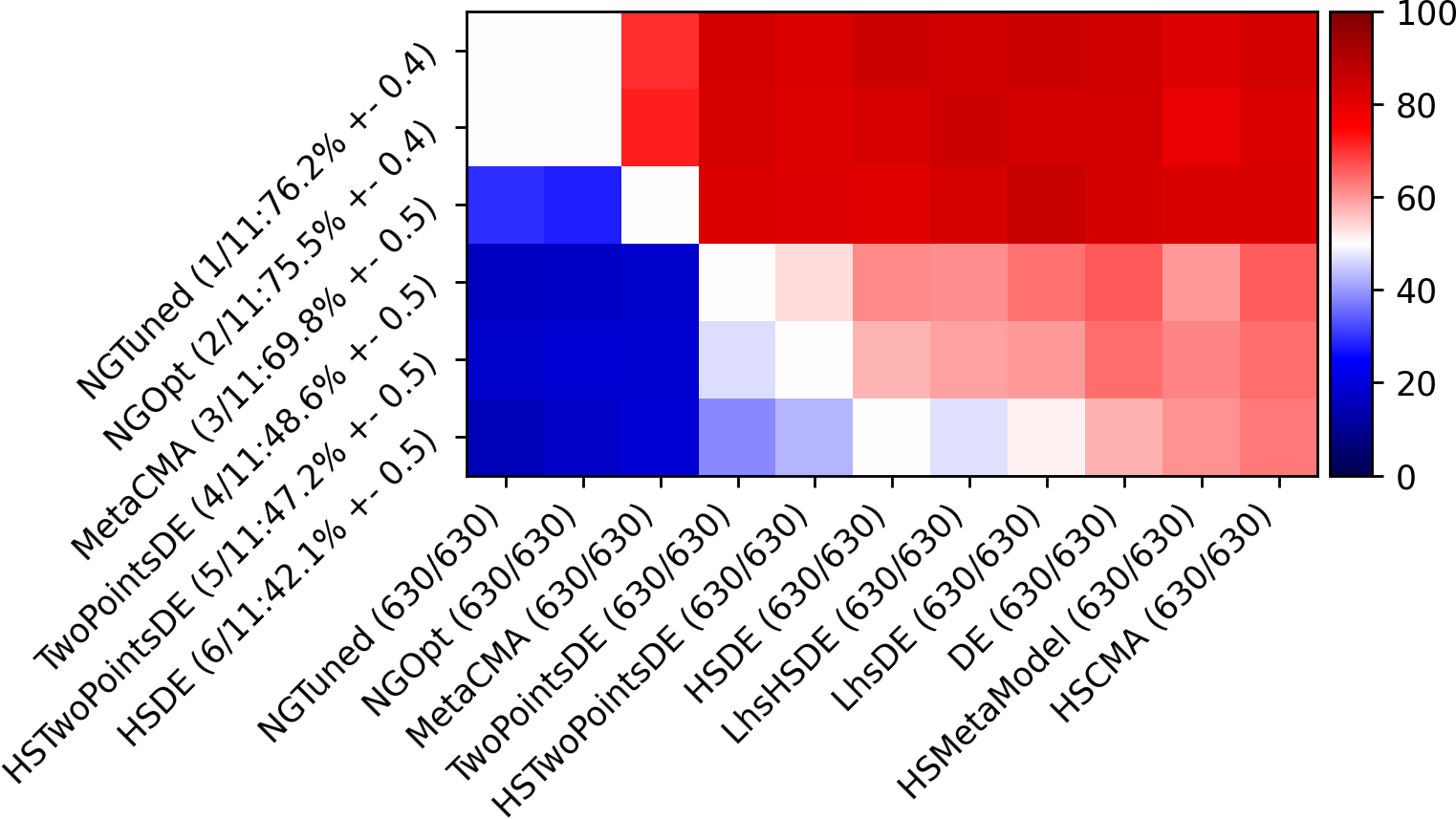}}
\includegraphics[height=3cm,width=.49\textwidth]{{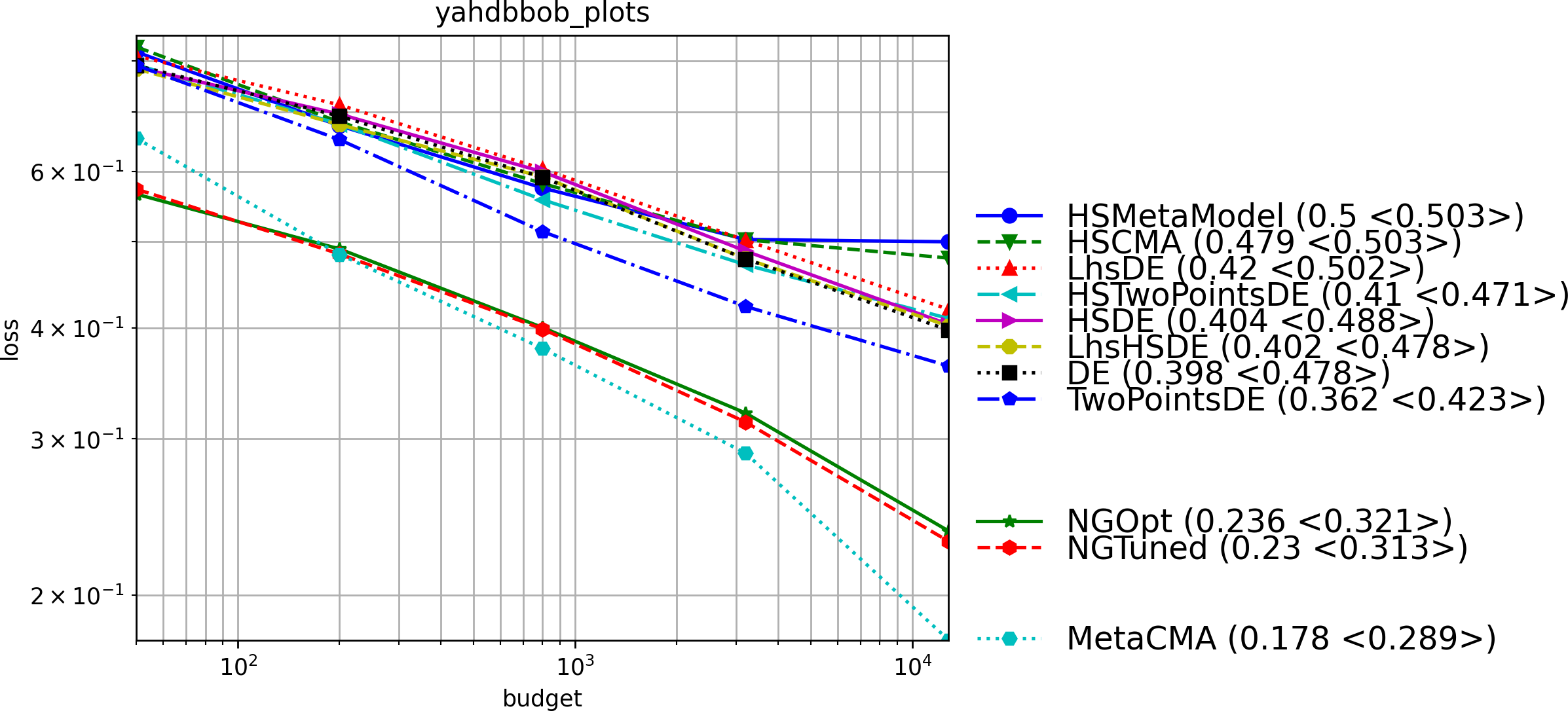}}\caption{\YAHDBBOB: high-dim. counterpart of \YABBOB, not used for training.}\label{yahdbbob}\end{figure}
\begin{figure}[!t]\centering\includegraphics[height=3cm,width=.49\textwidth]{{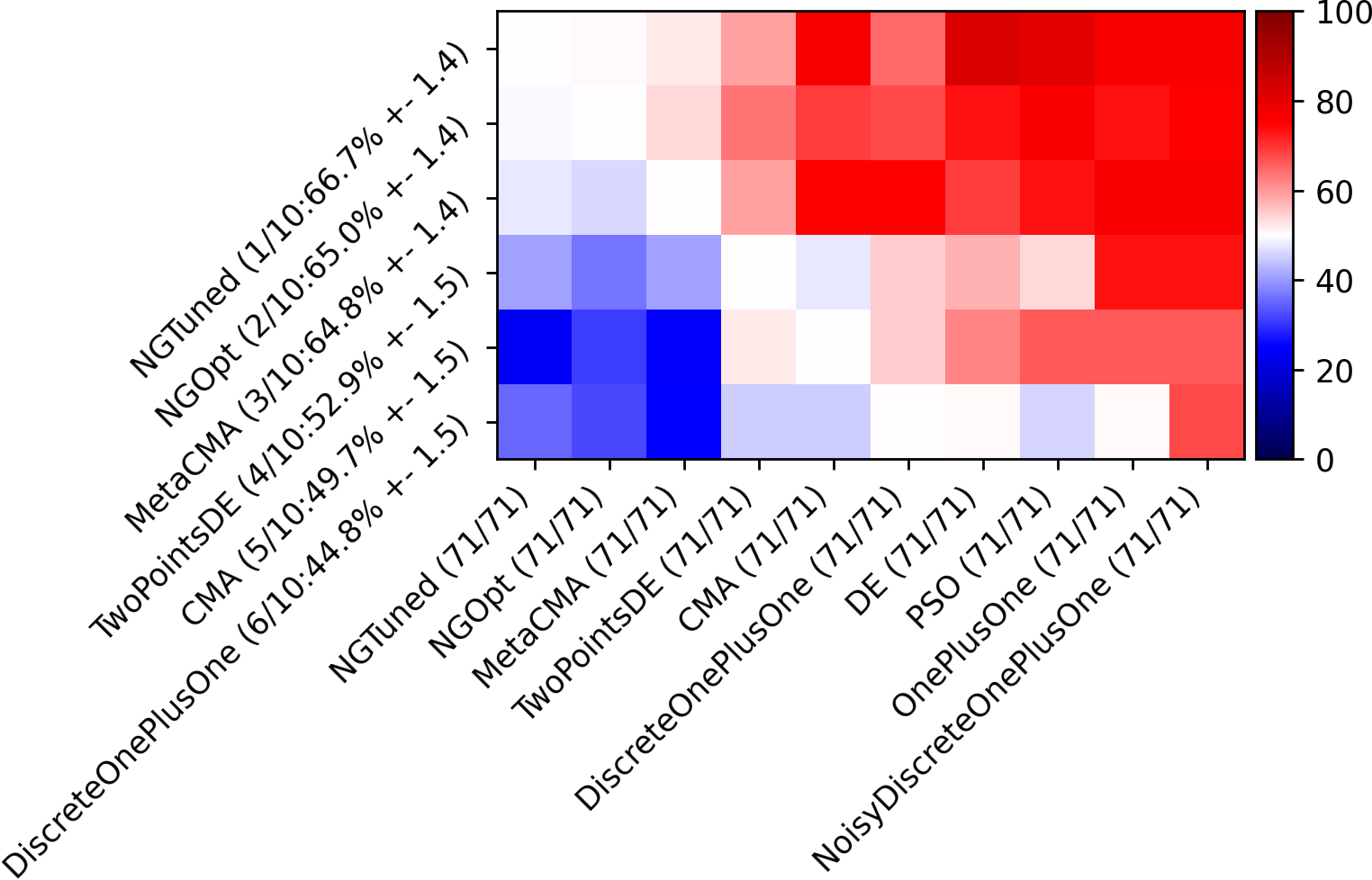}}
\includegraphics[height=3cm,width=.49\textwidth]{{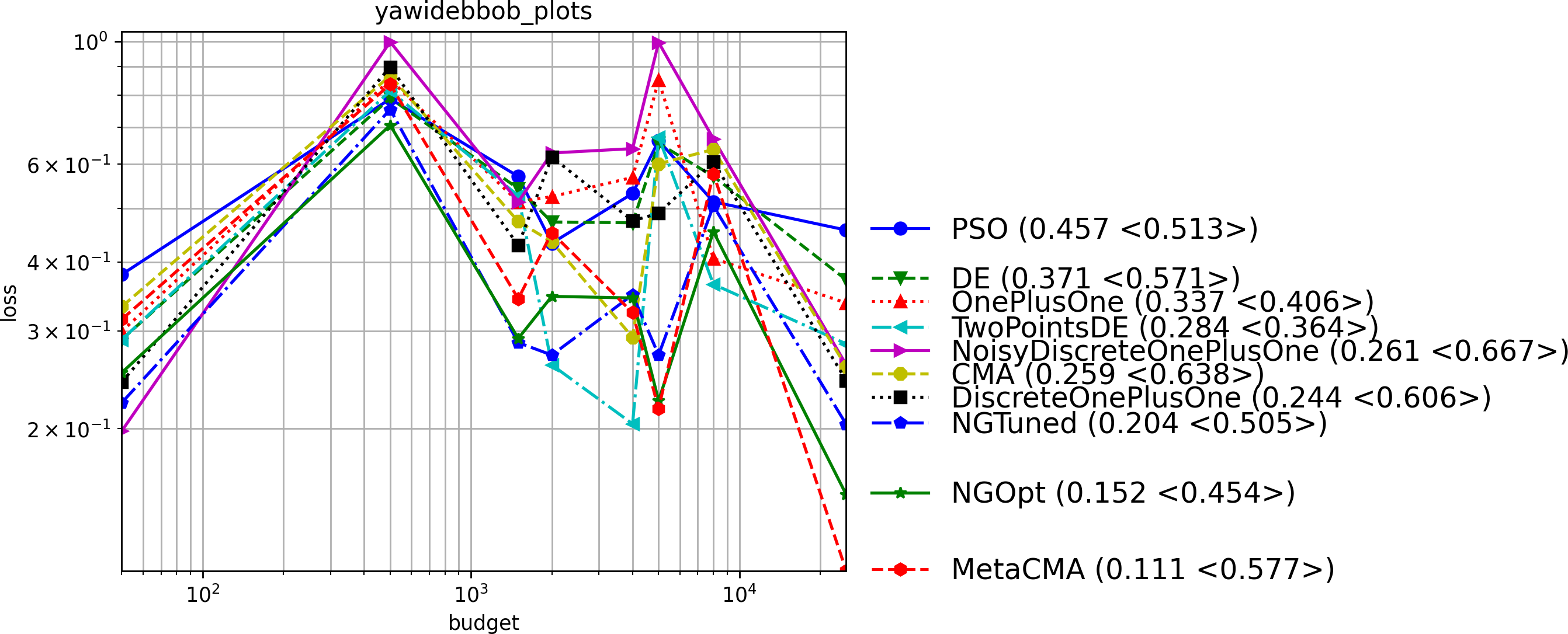}}\caption{Results on \YAWIDEBBOB: this benchmark matters particularly as it covers many different test cases, continuous, discrete, noisy, noise-free, parallel or not, multi-objective or not. It is designed for validating algorithm selection wizards. The different problems have different budgets, so that it is not surprising that the curve is not decreasing.}\label{yawidebbob}\end{figure}
\begin{figure}
\includegraphics[height=3.5cm,width=.49\textwidth]{{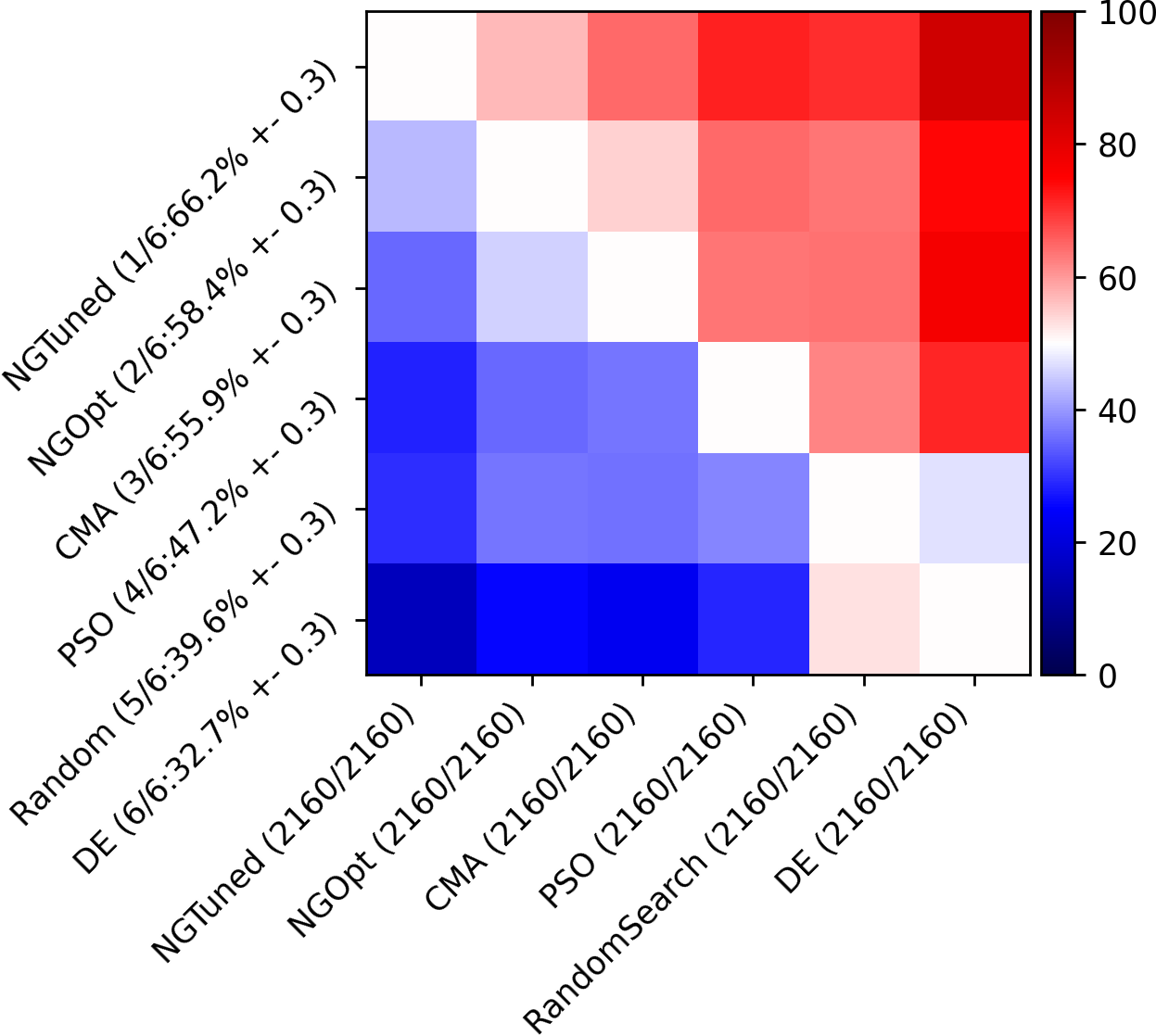}} 
\includegraphics[height=3.5cm,width=.49\textwidth]{{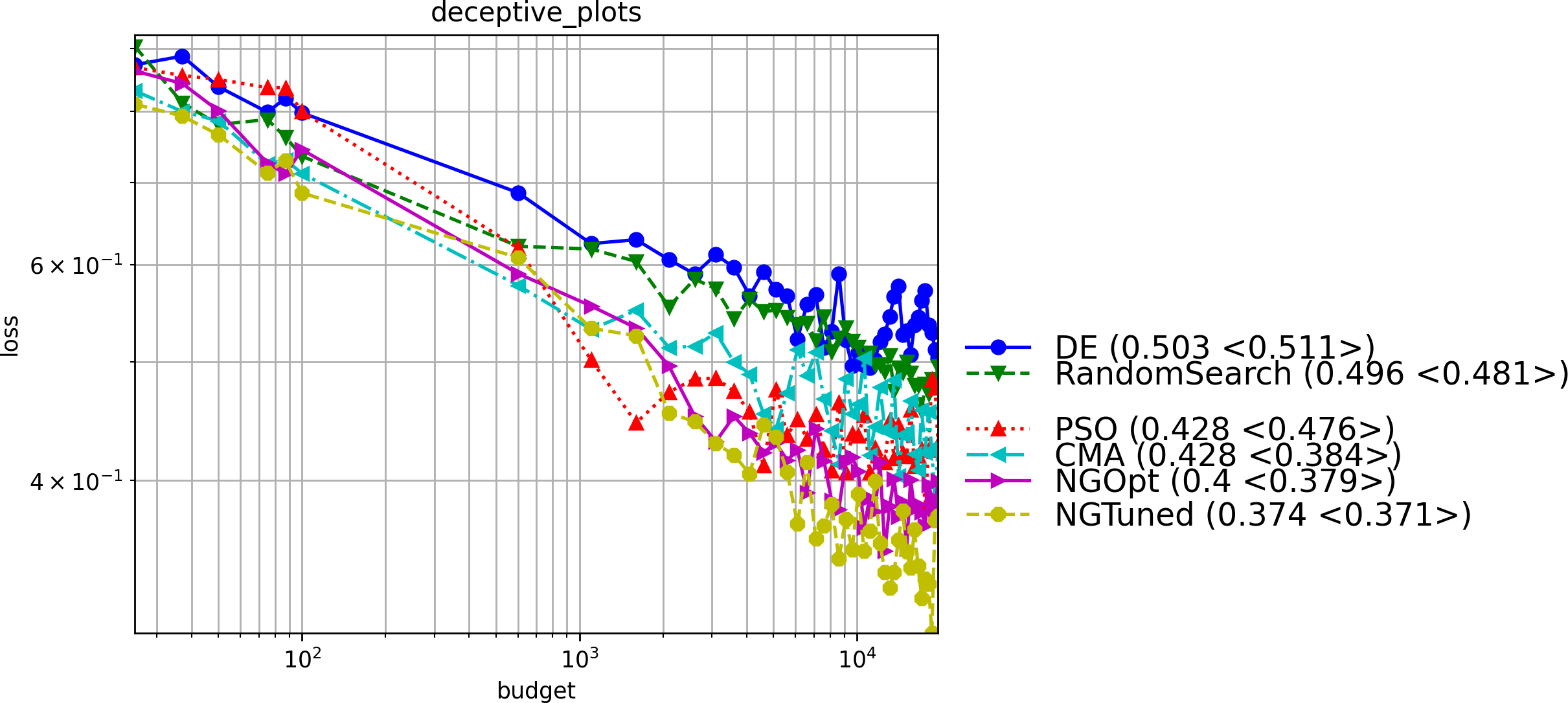}}
\caption{Deceptive  functions: this benchmark \cite{nevergrad} is quite orthogonal to the \textsc{ya*bbob} benchmarks, and very hard (infinitely many local minima arbitrarily close to the optimum / infinite condition number / infinitely tiny path to the optimum) .}
\label{deceptive}
\end{figure}}

\subsection{Evaluation on Benchmark Suites Not Used for Tuning}
\textbf{Artificial Benchmarks.}  Figures~\ref{yabigbbob},~\ref{yaboxbbob},~\ref{yawidebbob},~\ref{yahdbbob}, and~\ref{deceptive} present results on benchmark suites related to \YABBOB though with different configurations. The problem instances used were not used in the tuning process.

Figure~\ref{yabigbbob} shows the results  on the \YABIGBBOB benchmark suite, which consists of the same problem instances as the \YABBOB benchmark suite with longer budgets. Here, NGOpt achieves the best performance, while NGTuned is second ranked. From the line plot on the right of Figure~\ref{yabigbbob}, we can see that NGOpt and NGTuned are better than the other algorithms over all budgets.
The results for the \YABOXBBOB benchmark suite (box-constrained and low budget) are presented in Figure~\ref{yaboxbbob}. In this case, all of the proposed algorithms appear as top ranked, with MetaCMA being the best performing one. As for budget up to around $10^3$, MetaCMA performs the best, and is later outperformed by NGTuned, which uses the MetaModel on top of MetaCMA. 
The \YAHDBBOB benchmark suite is a high-dimensional counter part of \YABBOB. On it, NGTuned is selected as the best algorithm in Figure~\ref{yahdbbob}, followed by NGOpt with 0.7\% difference referring to the \% of times they outperform other algorithms. The right subplot shows that these algorithms together with MetaCMA achieved the best losses, across all budgets.

Figure~\ref{yawidebbob} shows results on the \YAWIDEBBOB benchmark suite, which is considered to be specially difficult and  covers many different test problem instances (continuous, discrete, noisy, noise-free, parallel or not, multi-objective or not). On \YAWIDEBBOB, NGTuned is selected as the best algorithm, followed by NGOpt, with a difference of 1.7\% in the average score related to the number of times they outperform the other algorithms. MetaCMA is ranked as third. From the line plot we can see that these algorithms achieved lowest losses, however the behavior is not stable over different budgets. We need to point out here that NGTuned is not using CMA for multi-objective problems, so that some parts of \YAWIDEBBOB are not modified by our work. The results indicate that the automated configuration that is performed for such benchmark suites also helps to improve the performance of the NGOpt on benchmark suites that contain a mix of different problem instances. %
The results for the Deceptive functions benchmark suite are presented in Figure~\ref{deceptive}, where NGTuned is the winning algorithm in terms of the number of times it performs better than the other algorithms and the mean loss achieved for different budgets over all settings. This benchmark suite is quite orthogonal to the \YABBOB variants and very hard to optimize, since the deceptive functions contain infinitely many local minima arbitrarily close to the optimum or have an infinite condition number or have an infinite tiny path to the optimum.

\textbf{Real-world Problems.} 
%
We also tested our algorithms on some of the real-world benchmark suites presented in Nevergrad.  Due to length constraints, the figures displaying these results are available only in the full version of this paper. 
On the SimpleTsp benchmark suite, NGTuned outperforms NGOpt in terms of the average frequency of winning over other algorithms and mean loss achieved for different budgets over all settings. 
On the SeqMlTuning benchmark NGTuned appears among the best-performing algorithms with rank six.
It outperforms NGOpt and MetaCMA which do not appear in the set of best performing algorithms.
On the ComplexTSP problem, NGTuned is found to be the best performing algorithm. It also surpasses NGOpt. 
With regard to mean performance per budget, they are very similar and consistently better than the other algorithms.
On the Unit Commitment benchmark suite, 
NGTuned is the third ranked algorithm. It outperforms NGOpt but not enough for competing with the best variants of DE. According to the convergence analysis, it seems that NGTuned and NGOpt have similar behavior.
On the Photonics problem,  
NGOpt and NGTuned appear between the top ranked algorithms showing very similar performance.
On th benchmark suite (GP), which is a challenge in genetic programming, NGOpt and NGTuned show almost identical but very bad performance overall: it is mentioned in~\cite{nevergrad} that NGOpt has counterparts dedicated to neural reinforcement learning (MixDeterministic RL and ProgNoisyRL) that perform better: we confirm this. 
On the 007 benchmark suite, 
NGTuned is the best performing algorithm. Regarding the mean loss achieved for different budgets, NGTuned shows similar performance to NGOpt.

\section{Conclusion}\label{sec:conclusion}
We have shown in this paper how automated algorithm configuration methods like irace may be used to improve an algorithm selection wizard like NGOpt. NGTuned was better than NGOpt in most benchmark suites, included those not used for tuning our MetaCMA.
    Our results improve the algorithm selection wizard not only for the benchmark suites used during the tuning process but also for a wider set of benchmark suites, including real-world test cases. Given the complexity of NGOpt, we have an impact on few cases, so that the overall impact on NGOpt across all Nevergrad suites remains moderate: in some cases, NGTuned is equivalent to NGOpt, and in some cases (in which CMA is crucial) it performs clearly better. In some cases, the differences between the two algorithms are small. If we consider only clear differences, we have gaps for YABBOB, 007,  YASMALLBBOB, YAPARABBOB, YABOXBBOB, SimpleTSP, and Deceptive: in all these cases, NGTuned performs better than NGOpt.
    
    Our tuning process still contains a few manual steps that require an expert, such as deciding which benchmark suites to use for tuning, as well as the integration of the tuned CMA configurations into NGOpt. Our ultimate goal is to automatize all steps of the process, and to also tune other base components as well as the decision rules that underlie the NGOpt algorithm selection process. 
    %
    %
\begin{smaller}[1]
\noindent%
\end{smaller}

\textbf{Acknowledgments.} M.\@ L\'opez-Ib\'a\~nez is a ``Beatriz Galindo'' Senior Distinguished Researcher (BEAGAL 18/00053) funded by the Spanish Ministry of Science and Innovation (MICINN). C. Doerr is supported by the Paris Ile-de-France Region (AlgoSelect) and by the INS2I institute of CNRS (RandSearch). T. Eftimov, A. Nikolikj, and G. Cenikj is supported by the Slovenian Research Agency: research  core  fundings  No.  P2-0098 and  project  No. N2-0239. G. Cenikj is also supported by the Ad Futura grant for postgraduate study.

\renewcommand{\doi}[1]{doi:\hspace{.16667em plus .08333em}\discretionary{}{}{}\href{https://doi.org/#1}{\urlstyle{rm}\nolinkurl{#1}}}
\FloatBarrier
\providecommand{\MaxMinAntSystem}{{$\cal MAX$--$\cal MIN$} {Ant} {System}}
  \providecommand{\rpackage}[1]{{#1}}
  \providecommand{\softwarepackage}[1]{{#1}}
  \providecommand{\proglang}[1]{{#1}}

\end{document}